\documentclass[conference]{IEEEtran}
\IEEEoverridecommandlockouts
\usepackage{cite}
\usepackage{amsmath,amssymb,amsfonts}
\usepackage{algorithmic}
\usepackage{algorithm}
\usepackage{graphicx}
\usepackage{textcomp}
\usepackage{xcolor}
\usepackage{comment}
\usepackage{caption}
\usepackage{makecell} 
\usepackage{multirow}
\usepackage{flushend}
\usepackage{mathtools}
\usepackage{enumitem}
\usepackage[hyphens]{url}
\usepackage[breaklinks=true]{hyperref}
\usepackage{subfiles}
\usepackage{booktabs}
\usepackage{tikz} 
\usepackage{pifont}
\usepackage{balance}

\graphicspath{{Images/}}
\def\BibTeX{{\rm B\kern-.05em{\sc i\kern-.025em b}\kern-.08em
    T\kern-.1667em\lower.7ex\hbox{E}\kern-.125emX}}

\begin{document}
\bstctlcite{IEEEexample:BSTcontrol}

\title{Contextual Bandits for Resource-Constrained Devices using Probabilistic Learning
\thanks{The work of AL and DK was supported by Knut and Alice Wallenberg Foundation under the Wallenberg Scholars program (Grant KAW2023.0327).
DK acknowledges funding from the Swedish Strategic Research Foundation under the Future Research Leaders program (Grant FFL24-0111), the Swedish Research Council under the Starting Grant program (Grant 2025-05421), and the Swedish Knowledge Foundation (Contract 20210016).}
}

\author{
\IEEEauthorblockN{Marco Angioli}
\IEEEauthorblockA{
\textit{Sapienza University} \\ 
\textit{of Rome}
}
\and
\IEEEauthorblockN{Kevin Johansson}
\IEEEauthorblockA{
\textit{Örebro University}
}
\and
\IEEEauthorblockN{Antonello Rosato}
\IEEEauthorblockA{
\textit{Sapienza University} \\ 
\textit{of Rome}
}
\and
\IEEEauthorblockN{Amy Loutfi}
\IEEEauthorblockA{
\textit{Örebro University} \\ 
\textit{Linköping University} \\ 
}
\and
\IEEEauthorblockN{Denis Kleyko}
\IEEEauthorblockA{
\textit{Örebro University} \\ 
\textit{Research Institutes of Sweden} \\ 
}
}

\maketitle

\begin{abstract}
Contextual bandits (CB) are online sequential decision-making problems under partial feedback that underpin many adaptive services. 
There is a growing demand to deploy CB agents directly on-device, under strict constraints on memory, compute, and energy. 
However, standard linear CB algorithms are often impractical for resource-constrained devices with their unfavorable scaling in computational and memory costs.
Recently, HD-CB, a CB approach based on hyperdimensional computing principles, has been proposed to model and solve CB problems by moving into high-dimensional spaces.
HD-CB offers faster convergence, favorable scalability, and improves memory efficiency compared to linear CB algorithms. 
However, its learning rule is accumulation-based: the values of action vectors grow over time, requiring high precision. 
While periodic binarization can prevent overflow in low-precision components, it may discard important information about magnitudes and degrade decision quality.
This paper introduces probabilistic HD-CB, a low-precision variant that replaces deterministic accumulation with a probabilistic update rule. 
At each step, only a random subset of vector components is updated, with a time-decaying update probability, and component values are constrained to a predefined range $[-\kappa,+\kappa]$. This approach enables low-precision components, prevents overflow without periodic binarization, and reduces the expected update cost in proportion to the fraction of updated components. Off-policy evaluation on standardized synthetic CB benchmarks using the Open Bandit Pipeline shows that probabilistic HD-CB consistently outperforms binarized HD-CB at equal precision, while approaching the performance of HD-CB with as few as 3 bits per component. 
\end{abstract}


\section{Introduction}
Contextual multi-armed bandits (CB) formalize sequential decision-making processes in which an agent repeatedly selects an action based on side information (the context) and observes only the reward of the chosen action. This setting captures the exploration-exploitation trade-off that arises in many interactive systems, including online advertising~\cite{li2010contextual}, personalized recommendations~\cite{Netflix, MICROSOFT, NYT, GAME}, hardware reconfiguration~\cite{angioli_automatic}, edge computing~\cite{EDGE1, EDGE2}, and clinical decision support~\cite{MD1}, where the agent must adapt while limiting negative impact on the user~\cite{li2010contextual}. 
CB can be seen as a particular case of reinforcement learning, where the selected action does not affect the next context, and the agent aims to maximize the immediate rewards after each action selection. 

Deploying CB agents on \emph{resource-constrained} devices remains challenging, despite a demand to move decision-making from cloud to resource-constrained devices. The issue is that standard linear CB algorithms have to maintain covariance matrices whose memory footprint scales quadratically with the dimensionality of the context features and whose updates require nontrivial linear-algebra kernels that can not be easily parallelized~\cite{LINUCBOPT}. These operations are not only computationally expensive, but also memory- and bandwidth-intensive, often conflicting with resource constraints where energy and on-chip memory are the primary bottlenecks.

It was proposed to alleviate these problems of standard linear CB algorithms using hyperdimensional computing, also known as vector symbolic architectures (HD/VSA)~\cite{Kanerva2009, kleyko_review_I}. HD/VSA was applied to the CB setting, resulting in an HD-CB approach~\cite{angioli2025hd} that models and solves the entire sequential decision-making process in high-dimensional spaces, using lightweight component-wise operations. This approach can be interpreted as a shallow randomized neural network~\cite{scardapane2017randomness}, in which the early (frozen) layers encode context vectors, while the trainable output layer learns high-dimensional representations of actions through associative learning rules~\cite{shen2023reducing}.

HD-CB achieves linear time complexity, faster convergence, improved scalability, and reduced memory footprint, making it suitable for hardware acceleration and deployment on resource-constrained devices.
However, the standard HD-CB update is \emph{accumulation-based}: vectors representing actions are incremented or decremented at every action selection, and the growth of their components is not limited over time. Preventing overflow, requires either numeric types with high precision, causing high computational and memory costs.

To address this problem and reduce these costs, the study in~\cite{angioli2025hdbin} introduced HD-CB\textsubscript{BIN}, a low-precision variant that uses vectors with $Q$-bit integer components and periodically applies binarization. During action selection, vectors are binarized to enable Hamming-distance similarity, and every $Q$ updates the accumulated values are discretized back to the binary domain to keep the components bounded.
However, while effective in reducing resource usage, this strategy relies on a \emph{deterministic, hard binarization} operation that may discard useful magnitude information accumulated in the vectors and lead to non-negligible performance degradation, especially in noisy or data-scarce regimes where retaining historical information with higher fidelity is beneficial.

This paper addresses the limitation of HD-CB from a different perspective by introducing a probabilistic variant of HD-CB, namely, HD-CB\textsubscript{PROB}. 
Instead of periodically binarizing action vectors, this approach \emph{bounds the learning by design to a predefined precision}, introducing a probabilistic update rule, which in turn is inspired by previous work on low-precision neural networks~\cite{appiah2009binary, kleyko2019integer}. 
The proposed HD-CB\textsubscript{PROB} approach updates only a randomly selected subset of vector components at each action selection, enabling the use of low-precision, saturated integer counters by controlling the learning rate with time-decaying update probabilities. 
This allows the agent to retain historical information with the high-fidelity in low-precision components. 
As a result, the overflow is avoided, update costs are decreasing over time, and learning no longer relies on periodic binarization steps that may severely affect the usefulness of learned representations. 

\section{Background} 
\label{sec:background}

\subsection{Contextual bandits}
\label{subsec:cb_background}
In a multi-armed CB setting, an agent interacts with an external environment over discrete time steps $t=1,\dots,T$. 
At each time step, the agent observes a context vector $\mathbf{x}_{t} \in \mathbb{R}^d$ (this notation is used even if contexts are action-dependent) describing the current state of the environment. 
Given a finite set of \( N \) possible actions, the agent selects an action \( a_t \), and receives a scalar reward \( r_{t,a_t} \), derived from an unknown probability distribution conditioned on the context and action. 
Rewards for the remaining $N-1$ actions $(a \neq a_t)$ are not observed. 
Agent's objective is to maximize the cumulative reward $\sum_{t=1}^T r_t$ over time.
CB algorithms' ability to adapt policies online from logged interactions makes them particularly suitable for a wide range of real-world systems~\cite{CB_review}. 
They are widely adopted in interactive systems such as online advertising and recommendation, clinical decision support, and adaptive resource management, where decisions must be made from partial feedback~\cite{CB_review}. 
Recent deployments further highlight the growing relevance of CB at the edge, where decisions must be made locally under tight memory and compute constraints~\cite{tewari17ads,dronebandits}. 
For example, CB have been deployed on mobile and wearable devices to personalize interventions~\cite{tewari17ads}. 
Similarly, CB has been used on resource-constrained aerial devices to adaptively offload inference between edge and cloud~\cite{dronebandits}. 
In the domain of assistive robotics, CB-based approach has been proposed to personalize robot-assisted feeding~\cite{banerjee2025ask}.
These scenarios emphasize the increasing need for CB algorithms that are more efficient, scalable, and computationally lightweight.

\subsection{Linear contextual bandits}
\label{subsec:linear_cb}
A widely adopted family of CB algorithms, which will be considered as a baseline for this paper, assumes a linear relationship between the reward and the context~\cite{li2010contextual, li2011unbiased}. 
In this formulation, each action $a$ is associated with an unknown weight vector $\boldsymbol{\theta}_a\in\mathbb{R}^d$ such that
\begin{equation}
\mathbb{E}[r_{t,a}\mid \mathbf{x}_{t}] = \mathbf{x}_{t}^\top \boldsymbol{\theta}_a,
\end{equation}
with sub-Gaussian noise.
$\boldsymbol{\theta}_a$ is estimated online via the ridge regression by maintaining per-action statistics: a covariance matrix $\mathbf{A}_a \in \mathbb{R}^{d\times d}$ and a bias vector $\mathbf{b}_a \in \mathbb{R}^{d}$.

At each time step, the agent observes the input context $\mathbf{x}_{t}$, estimates the expected rewards using the ridge regression, and selects the action with the highest estimated value: 
\newline
\centerline{$
\boldsymbol{\hat{\theta}}_a=\mathbf{A}_a^{-1} \boldsymbol{b}_a,\qquad a_t = \operatorname*{argmax}_{a\in[1, N]} \left(
\mathbf{x}_{t}^\top \boldsymbol{\hat{\theta}}_a\right).
$}
Upon taking the selected action $a_t$  and observing reward $r_{t,a_t}$, the statistics for $a_t$ are updated:
\newline
\centerline{$
        \mathbf{A}_{a_{t}} = \mathbf{A}_{a_{t}}+\mathbf{x}_{t} \mathbf{x}_{t}^{\top}, \qquad
        \mathbf{b}_{a_{t}} = \mathbf{b}_{a_{t}}+r_{t, a_t} \mathbf{x}_{t}.
$}

To balance exploration of less-known actions and exploitation of collected statistics, linear CB algorithms integrate exploration strategies, ensuring the agent improves over time. 
In this paper, we consider the simple $\varepsilon$-greedy exploration strategy. 
At each time step, the agent selects the action with the highest expected reward with probability $1-\varepsilon$; with probability $\varepsilon$, it instead selects an action uniformly at random.
We refer to this approach as \emph{LinEPS}.

While effective and widely adopted, linear CB algorithms incur a memory footprint that scales quadratically with $d$ ($\mathcal{O}(Nd^2)$), and their asymptotic computational complexity grows as \(\mathcal{O}(d^3)\). 
The Sherman-Morrison formula can reduce the computational complexity of matrix inversion from $\mathcal{O}(d^3)$ to $\mathcal{O}(d^2)$~\cite{LINUCBOPT}, but the quadratic scaling still remains a bottleneck for high-dimensional contexts, motivating alternatives with simpler operations and smaller memory footprints.

\subsection{Hyperdimensional computing}
\label{subsec:hdc_background}
HD/VSA is a computing paradigm inspired by the high-dimensional, distributed representation of information in neural circuits~\cite{Kanerva2009}.
Information in HD/VSA is encoded using high-dimensional distributed representations (hypervectors) that are manipulated using a set of simple vector operations: i) \emph{superposition}, which aggregates multiple items; ii) \emph{binding}, which associates items; and iii) \emph{permutation}, which encodes order and roles; and are compared using a similarity measure.
Together, these four operations form the ``arithmetic'' of HD/VSA, enabling compositional representations of complex structures \emph{without increasing dimensionality}.
The implementation of these operations varies depending on the characteristics of the vector space in which hypervectors are defined~\cite{schlegel2022comparison}; in this paper we will consider the multiply–add–permute (MAP)~\cite{map} model, where the binding is defined as element-wise multiplication, superposition as addition, permutation as cyclic shift, and similarity as the inner product.

The distributed nature of representations and the simplicity of its arithmetic endows HD/VSA with several attractive properties, including robustness to noise, energy and computational efficiency, and inherent parallelism, that have made HD/VSA a compelling and promising alternative for deployment in resource-constrained devices, digital hardware accelerators, and unconventional computing hardware~\cite{kleyko_pieee}.

\subsection{Hyperdimensional contextual bandits}
\label{subsec:hdcb_background}
The hyperdimensional CB (HD-CB) approach~\cite{angioli2025hd} leverages HD/VSA to model and solve CB problems by transforming them into a high-dimensional space.
Instead of maintaining covariance matrices, HD-CB maintains a single hypervector $\mathcal{A}_a$ for each action $a$.
At time step $t$, the context is encoded into an integer hypervector $\mathcal{X}_{t} \in \mathbb{Z}^D$. Actions are then ranked by computing the similarity between the current context hypervector $\mathcal{X}_{t}$ and all action hypervectors $\mathcal{A}_a$.

To select actions, HD-CB can be combined with exploration strategies~\cite{angioli2025hd}. As with LinEPS, the $\varepsilon$-greedy strategy is used to demonstrate the proposed approach and compare to other alternatives. With probability $1-\varepsilon$, the agent selects $a_t = \arg\max_a \big(\frac{\mathcal{A}_a \mathcal{X}_{t}}{\| \mathcal{A}_a \|_2 \| \mathcal{X}_{t}\|_2 }\big)$, and a random action otherwise.

After selecting $a_t$ and observing reward $r_{t,a_t}$, HD-CB encodes the reward into a hypervector $\mathcal{R}_{t,a_t}\in\{-1,+1\}^D$ and updates the hypervector of the chosen action via superposition:
\begin{equation}
    \label{eq:hdcb_unbounded}
    \mathcal{A}_{a_t} = \mathcal{A}_{a_t} + \mathcal{R}_{t,a_t} \odot \mathcal{X}_{t}.
\end{equation}
The reward hypervector is encoded via a thermometer encoding~\cite{angioli2025hd, Rachkovskiy_2005, DensityEncoding2021}). Instead of ridge regression, the approach uses hardware-friendly vector operations scaling as $\mathcal{O}(D)$ and achieves faster convergence than linear CB algorithms~\cite{angioli2025hd}.

\subsection{Binarized HD-CB}
In the update rule of HD-CB, Eq.~\eqref{eq:hdcb_unbounded}, the \emph{unlimited growth} of values of $\mathcal{A}_{a_t}$ is problematic for resource-constrained devices. 
Since one of the action hypervectors is updated at every time step, values of $\mathcal{A}_a$ increase with time, growing indefinitely as $t \to \infty$.
Thus, this approach (referred to as HD-CB\textsubscript{REAL}) requires components of action hypervectors with high-precision (e.g., 32-bit), leading to a large memory footprint that limits its efficiency gains.

To reduce the memory footprint, a binarized variant of HD-CB (HD-CB\textsubscript{BIN}) has been proposed~\cite{angioli2025hdbin}. 
It stores action hypervectors using a $Q$-bit integers and leverages two distinct binarization mechanisms to control inference and overflow.

We explain now how this can be achieved practically.
First, temporary, binarized versions of action hypervectors are computed via a majority rule operation~\cite{kanerva1995family}, and by maintaining auxiliary counters for each action hypervector. 
The use of binarized versions enables efficient action selection as Hamming distance can be used for a similarity measure.
Second, to prevent overflow and keep the representation bounded, the hypervectors are periodically discretized back to the binary domain. 
Specifically, after every $2^Q$ updates, the components of a given hypervector $\mathcal{A}_a$ are replaced by their binarized counterparts, and the per-action counter is reset.

While reducing computational and memory costs, these mechanisms rely on \emph{hard binarization}. 
The periodic reset indiscriminately discards the information accumulated in counters between resets. 
A counter that has a value of, e.g., $100$ becomes indistinguishable from a counter with $2$. 
This raises a design choice between HD-CB, with its large memory footprint, and HD-CB\textsubscript{BIN} with its performance degradation after the periodic resets. 
As an alternative, this paper proposes an approach where action hypervectors are bounded-by-design.

\subsection{Learning with low-precision integer components} \label{subsec:isom_background}
The probabilistic update rule leveraged in this paper is inspired by the previous studies of self-organizing maps that proposed a reformulation of the algorithm, designed to enable low-cost learning in digital hardware~\cite{appiah2009binary,kleyko2019integer}.
While this update rule was presented in another context, the formulation can be used for updating signed integer components via simple increment/decrement steps. 
Two concepts are relevant to the CB setting: i) a bounded integer state enforced by saturation and ii) a probabilistic per-component update controlled by an update probability that decreases over time~\cite{appiah2009binary,kleyko2019integer}.

Unlike the standard accumulation-based update, which allows hypervector components to grow indefinitely, the probabilistic update rule constrains each component to a predefined integer range $[-\kappa,+\kappa]$. This is achieved by applying a clipping function $f_{\kappa}(\cdot)$ after every update:
\newline
\centerline{$
f_{\kappa}(v)= \max(-\kappa, \min(v, +\kappa)),
$}
where $v$ denotes the value being clipped, and $\kappa$ is a configurable saturation threshold. It controls the trade-off between representational capacity and precision: small $\kappa$ yields very low-precision components (e.g., ternary for $\kappa=1$), while larger $\kappa$ allows the agent to accumulate more information.
If $v$ is restricted to integers, $f_{\kappa}(v)$ can be stored within $\lceil \log_2(2\kappa+1) \rceil$ bitwidth, preventing the unlimited growth typical of standard accumulation-based updates. 

To enable learning within such low-precision components, the deterministic accumulation is replaced with \emph{probabilistic} updates. Let $w_{t,i}$ denote the $i$-th component of the weight vector at time step $t$, and let $\mathcal{T}_{t,i} \in \{-1,+1\}$ denote the corresponding component of the current update signal. Instead of using $\mathcal{T}_{t,i}$ at every update, components are updated with probability $\alpha_t$. For each component $i$, a random value $m_{t,i} \sim \mathcal{U}(0,1)$ is sampled; the update is performed only if $m_{t,i} \le \alpha_t$:
\begin{equation}
w_{t,i}=
\begin{cases}
f_{\kappa}\!\big(w_{t-1,i}+\mathcal{T}_{t,i}\big), & \text{if } m_{t,i} \le \alpha_t,\\
w_{t-1,i}, & \text{otherwise}.
\end{cases}
\label{eq:isom_update}
\end{equation}

In contrast to other learning rules scaling the update value over time, the rule in Eq.~\eqref{eq:isom_update} scales \emph{frequency} of updates via $\alpha_t$. By adopting this principle, the action hypervectors can be updated using only integer addition and saturation, achieving a low-precision integer components and reduced write-access frequency without the information loss of binarization.

\begin{algorithm}[t]
\caption{Probabilistic HD-CB with $\varepsilon$-greedy strategy}
\label{alg:prob_hdcb}
\begin{algorithmic}[1]
\STATE \textbf{Input:} $N$, $T$, $D$, $\varepsilon$, $\kappa$, $\alpha_0$, 
\STATE \textbf{Initialization:} Set $\mathcal{A}_a$ to $\mathbf{0}^D$ for all $a \in \{1,\dots,N\}$
\FOR{$t = 1, \dots, T$}
    \STATE \textbf{// 1. Encoding context $\mathbf{x}_{t}$  }
    \STATE Encode $\mathbf{x}_{t} \to \mathcal{X}_{t} \in \{-1, +1\}^D$
    
    \STATE \textbf{// 2. Action selection ($\varepsilon$-greedy strategy)}
    \STATE Draw $q \sim \mathcal{U}(0,1)$
    \IF{$q < \varepsilon$}
        \STATE Select random action $a_t \sim \mathcal{U}\{1, \dots, N\}$
    \ELSE
        \STATE Select $a_t = \operatorname*{arg\,max}_{a} \left( \mathcal{X}_{t}^\top \mathcal{A}_a \right)$
    \ENDIF
    
    \STATE \textbf{// 3. Observation}
    \STATE Execute action $a_t$, observe reward $r_{t,a_t}$
    
    \STATE \textbf{// 4. Probabilistic update rule}
    \STATE Compute decay $\alpha_t \leftarrow \alpha_0 \max(0, 1 - \frac{t-1}{T})$
    \STATE Encode $\mathbf{r}_{t,a_t} \to \mathcal{R}_{t,a_t} \in \{-1, +1\}^D$
    \STATE Compute update hypervector $\mathcal{T}_t \leftarrow \mathcal{R}_{t,a_t} \odot \mathcal{X}_{t}$
    
    \FOR{$i = 1, \dots, D$}
        \STATE Draw $\mathcal{M}_{t,i} \sim \mathcal{U}(0,1)$
        \IF{$\mathcal{M}_{t,i} < \alpha_t$}
            \STATE Updated value  $v \leftarrow \mathcal{A}_{a_t, i} + \mathcal{T}_{t, i}$
            \STATE Saturated value  $\mathcal{A}_{a_t, i} \leftarrow \max(-\kappa, \min(v, +\kappa))$ 
        \ENDIF
    \ENDFOR
\ENDFOR
\end{algorithmic}
\end{algorithm}

\section{Probabilistic HD-CB}
\label{sec:method}
This section introduces HD-CB\textsubscript{PROB}, a variant of HD-CB designed to operate under strict memory constraints without the information loss associated with periodic resets as in HD-CB\textsubscript{BIN}. 
The core intuition is to replace the unlimited, accumulation-based update of the standard HD-CB with the probabilistic update rule on components with low-precision. 
By treating the learning rate not as a scaling factor for the update value, but as a \emph{probability of update}, we can design an agent using components with low-precision saturating integers.

\subsection{Algorithm overview}
HD-CB\textsubscript{PROB} (Algorithm~\ref{alg:prob_hdcb}) maintains a set of action hypervectors $\mathcal{A}_a \in [-\kappa, +\kappa]^D$, one per action $a \in \{1, \dots, N\}$. 
Unlike standard HD-CB, where components are high-precision integers or floats, here every component $\mathcal{A}_{a,i}$ is strictly bounded by the saturation threshold $\kappa$.
It proceeds as follows:
\begin{enumerate}
    \item \textit{Context encoding:} Context $\mathbf{x}_{t}$ is encoded into a hypervector $\mathcal{X}_{t} \in \{-1, +1\}^D$ using a record-based transformation~\cite{angioli2025hd}. The fact that it is bipolar is important, as updates are done simply in $\pm 1$ steps, remaining compatible with low-precision saturating integers.
    \item \textit{Action selection:} The agent computes the similarity between the context $\mathcal{X}_{t}$ and all action hypervectors $\mathcal{A}_a$, selecting action $a_t$ via an $\varepsilon$-greedy strategy. Notably, in this variant, similarity can be efficiently computed using the inner product, as the norm of the saturating hypervectors remains implicitly bounded over time.
    \item \textit{Probabilistic update:} Reward $r_{t,a_t}$ is encoded into a hypervector $\mathcal{R}_{t,a_t}$ (using thermometer encoding) and bound with $\mathcal{X}_{t}$ to form an update hypervector. Instead of adding it to $\mathcal{A}_a$, the probabilistic update rule is applied, randomly selecting a subset of components being modified, and the result is clipped to the range $[-\kappa, \kappa]$.
\end{enumerate}

\subsection{Probabilistic update rule}
The key feature of HD-CB\textsubscript{PROB} is its update rule. Following the probabilistic update rule in Eq.~\eqref{eq:isom_update}, each component is updated with probability $\alpha_t$ and then clipped to the range $[-\kappa, \kappa]$.
The parameter $\alpha_t \in [0, 1]$ reflects a time-decaying learning schedule and is computed as $\alpha_t = \alpha_0 \left(1 - (t-1)/T \right)$ for $t=1,\dots,T$, where $\alpha_t=0$ for $t > T$.

At each time step, for each component $i \in \{1, \dots, D\}$, an i.i.d. random variable $m_{t,i} \sim \mathcal{U}(0, 1)$ is sampled and a binary update mask $\mathcal{M}_{t,i} \in \{0, 1\}$ is computed as 
$\mathcal{M}_{t,i} = \mathbb{I}(m_{t,i} \le \alpha_t)$,
where $\mathbb{I}(\cdot)$ is the indicator function.
The update directions are determined by the result of the binding of the reward and context hypervectors:
$\mathcal{T}_{t} = \mathcal{R}_{t,a_t} \odot \mathcal{X}_{t} \in \{-1, +1\}^D$.
Finally, the chosen action hypervector is updated as:
\begin{equation}
\label{eq:prob_update}
\mathcal{A}_{a_t} = f_{\kappa} \left( A_{a_t} + \mathcal{M}_{t} \odot \mathcal{T}_{t} \right).
\end{equation}
Intuitively, this update acts as a \emph{saturating random walk} driven by the bipolar update signal: active components (where $\mathcal{M}_{t,i}=1$) take a single $\pm 1$ step towards the saturation boundaries $\pm\kappa$, ensuring overflow is always prevented while preserving the direction of the accumulated hypervectors.

HD-CB\textsubscript{PROB} addresses the limitations of periodic binarization while preserving the properties that motivated HD-CB\textsubscript{BIN}.
\textit{Low-precision components}, $\mathcal{A}\in[-\kappa,\kappa]^{D \times N}$, are guaranteed by design via saturation.
\textit{Update cost is reduced} since updates occur only when $\mathcal{M}_{t,i} = 1$ and the number of write operations to memory is controlled by $\alpha_t$. 
In later stages when $\alpha_t$ is getting close to zero, the updates become highly sparse, significantly reducing computational costs.
\textit{Periodic resets are not needed} as the agent can accumulate information in $\mathcal{A}\in[-\kappa,\kappa]^{D \times N}$ without periodically binarizing to $\{-1,+1\}$.

\begin{table*}[!t]
    \caption{Off-policy evaluation results on the OBP library. The table compares the real-valued variant against two variants with low-precision components. 
    Best values among these variants are underlined; global bests are in bold. Results are averaged over 50 independently generated datasets; $\pm$ values denote standard deviation across datasets.}
    \label{tab:hdcb_eps_variants}
    \centering
    \resizebox{\textwidth}{!}{%
    \begin{tabular}{cc | c c | ccc | ccc}
        \toprule
        \multirow{2}{*}{\textbf{\textit{N}}} & \multirow{2}{*}{\textbf{\textit{d}}} &
        \multirow{2}{*}{\textbf{LinEPS}} &
        \multirow{2}{*}{\textbf{HD-CB\textsubscript{REAL}}} &
        \multicolumn{3}{c|}{\textbf{HD-CB\textsubscript{BIN}}} &
        \multicolumn{3}{c}{\textbf{HD-CB\textsubscript{PROB} (Proposed)}} \\
        & & & &
        \textbf{2 bits} & \textbf{3 bits} & \textbf{4 bits} &
        \textbf{2 bits} & \textbf{3 bits} & \textbf{4 bits} \\
        \midrule
        10 & 5  & 0.665$\pm$.076 & \textbf{0.687}$\pm$.073 & 0.606$\pm$.076 & 0.620$\pm$.079 & 0.630$\pm$.079 & 0.659$\pm$.084 & \underline{0.681}$\pm$.072 & 0.675$\pm$.079 \\
        10 & 10 & 0.683$\pm$.050 & \textbf{0.684}$\pm$.060 & 0.581$\pm$.055 & 0.598$\pm$.055 & 0.614$\pm$.060 & 0.631$\pm$.061 & \underline{0.652}$\pm$.059 & 0.635$\pm$.054 \\
        10 & 15 & \textbf{0.663}$\pm$.051 & 0.659$\pm$.063 & 0.569$\pm$.074 & 0.569$\pm$.069 & 0.588$\pm$.069 & 0.603$\pm$.069 & \underline{0.631}$\pm$.062 & 0.618$\pm$.059 \\
        15 & 5  & 0.711$\pm$.076 & \textbf{0.735}$\pm$.071 & 0.645$\pm$.084 & 0.663$\pm$.082 & 0.682$\pm$.078 & 0.707$\pm$.084 & \underline{0.718}$\pm$.069 & 0.705$\pm$.073 \\
        15 & 10 & 0.703$\pm$.049 & \textbf{0.708}$\pm$.050 & 0.598$\pm$.057 & 0.614$\pm$.058 & 0.636$\pm$.057 & 0.642$\pm$.054 & \underline{0.671}$\pm$.048 & 0.643$\pm$.045 \\
        15 & 15 & \textbf{0.678}$\pm$.067 & 0.664$\pm$.065 & 0.565$\pm$.073 & 0.573$\pm$.069 & 0.583$\pm$.070 & 0.601$\pm$.072 & \underline{0.630}$\pm$.062 & 0.608$\pm$.055 \\
        20 & 5  & 0.716$\pm$.078 & \textbf{0.733}$\pm$.071 & 0.643$\pm$.084 & 0.658$\pm$.084 & 0.677$\pm$.085 & 0.700$\pm$.083 & \underline{0.728}$\pm$.083 & 0.718$\pm$.080 \\
        20 & 10 & 0.708$\pm$.052 & \textbf{0.724}$\pm$.065 & 0.614$\pm$.076 & 0.634$\pm$.079 & 0.656$\pm$.076 & 0.661$\pm$.083 & \underline{0.685}$\pm$.067 & 0.657$\pm$.065 \\
        20 & 15 & \textbf{0.690}$\pm$.065 & 0.681$\pm$.071 & 0.584$\pm$.082 & 0.591$\pm$.079 & 0.601$\pm$.084 & 0.623$\pm$.080 & \underline{0.656}$\pm$.070 & 0.621$\pm$.061 \\
        \bottomrule
    \end{tabular}%
    }
\end{table*}

\section{Experimental Evaluation}
\label{sec:results}
To validate the proposed approach, we conducted an off-policy evaluation using the Open Bandit Pipeline (OBP) library~\cite{saito2020open} and performed a comparative analysis of the memory footprint. OBP provides a controlled, fully reproducible evaluation setting consistent with prior HD-CB works, enabling direct comparison under identical conditions.
These experiments quantify the trade-off between the low-precision action hypervectors formed with the probabilistic update rule and the quality of the decision-making.
We compare four approaches introduced earlier. 
To be consistent, all of them use the $\varepsilon$-greedy exploration strategy: 
\begin{enumerate}
    \item LinEPS: Standard linear CB algorithm based on covariance matrices and the ridge regression~\cite{li2010contextual}.
    \item HD-CB\textsubscript{REAL}: Real-valued HD-CB variant~\cite{angioli2025hd}.
    \item HD-CB\textsubscript{BIN}: Binarized variant, relying on periodic binarization to prevent overflow~\cite{angioli2025hdbin}.
    \item HD-CB\textsubscript{PROB}:\,Approach\,with probabilistic\,updates,\,Eq.~\eqref{eq:prob_update}.
\end{enumerate}

\begin{figure}[!b]
    \centering
    \includegraphics[width=1.0\columnwidth]{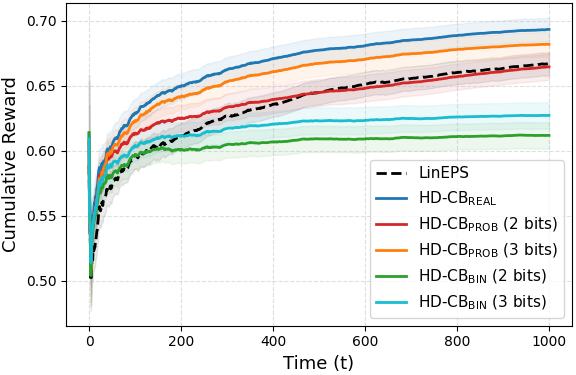}
    \caption{Cumulative rewards against the number of action selection rounds for six different agents. The results are averaged of $50$ datasets from  the OBP library, $d=5$, $N=10$, $D=1024$. For HD-CB\textsubscript{PROB} and HD-CB\textsubscript{BIN} variants at $2$ and $3$ bits per component are depicted. Shaded regions denote $\pm 1$ standard error of the mean across datasets.
    }
    \label{fig:OBP:reward}
\end{figure}

\subsection{Off-policy evaluation}
We first reports results obtained on the OBP library in the case of binary rewards.  
Cumulative rewards are averaged over 50 datasets per configuration. 
The dimensionality of hypervectors is fixed at $D=1024$. 
For HD-CB\textsubscript{PROB}, the initial update probability is set to $\alpha_0=0.4$.

Fig.~\ref{fig:OBP:reward} depicts the average cumulative rewards for the considered approaches against the number of action selection rounds when context dimensionality $d = 5$ and number of actions $N=10$ while Table~\ref{tab:hdcb_eps_variants} presents the results after $t=1000$ time steps for varying number of actions $N \in \{10, 15, 20\}$ and context dimensionality $d \in \{5, 10, 15\}$. 
For low-precision approaches, we evaluate configurations at $2$, $3$, and $4$ (not shown in Fig.~\ref{fig:OBP:reward}) bits per component. 
For HD-CB\textsubscript{PROB}, these correspond to saturation thresholds of $\kappa=1$ (ternary), $\kappa=3$, and $\kappa=7$, respectively. 
For HD-CB\textsubscript{BIN}, instead, these correspond to $Q = 2, 3$ and $4$, that is $2^Q = 4, 8, 16$, respectively.
The exploration parameter $\varepsilon$ is tuned via grid search for each configuration, and the best cumulative rewards on the OBP library are reported.

Three important observations could be made regarding the obtained numerical results.  
First, HD-CB\textsubscript{PROB} outperforms HD-CB\textsubscript{BIN} for every tested configuration $(N,d)$ and every bitwidth, it yields higher cumulative rewards than HD-CB\textsubscript{BIN} at any time step (cf. Fig.~\ref{fig:OBP:reward}). 
The advantage is particularly pronounced at low precision: averaged across all configurations, 3-bit HD-CB\textsubscript{PROB} ($\kappa=3$) improves over 3-bit HD-CB\textsubscript{BIN} by approximately $5.8\%$.
This confirms the hypothesis that adding information via probabilistic updates preserves more ranking information than periodic binarization.
Second, 3-bit HD-CB\textsubscript{PROB} consistently approaches the performance of HD-CB\textsubscript{REAL} baseline, with an average gap of only $2.5\%$. 
This indicates that even very low precision is sufficient to capture the relative quality of actions, provided the update rule permits the accumulation of relevant information.
Third, it is notable that increasing precision from $3$ to $4$ bits yields diminishing returns (or slightly negative) returns for HD-CB\textsubscript{PROB}. 
This suggests that, in this setting, increasing precision provides limited additional benefit once the agent can already store a modest amount of information per component. 
Additionally, the regularization effect of the tighter saturation bound $\kappa=3$ may actually be beneficial in these synthetic scenarios.

\subsection{Memory footprint analysis}

\begin{figure}[t!]
    \centering
    \includegraphics[width=0.92\linewidth]{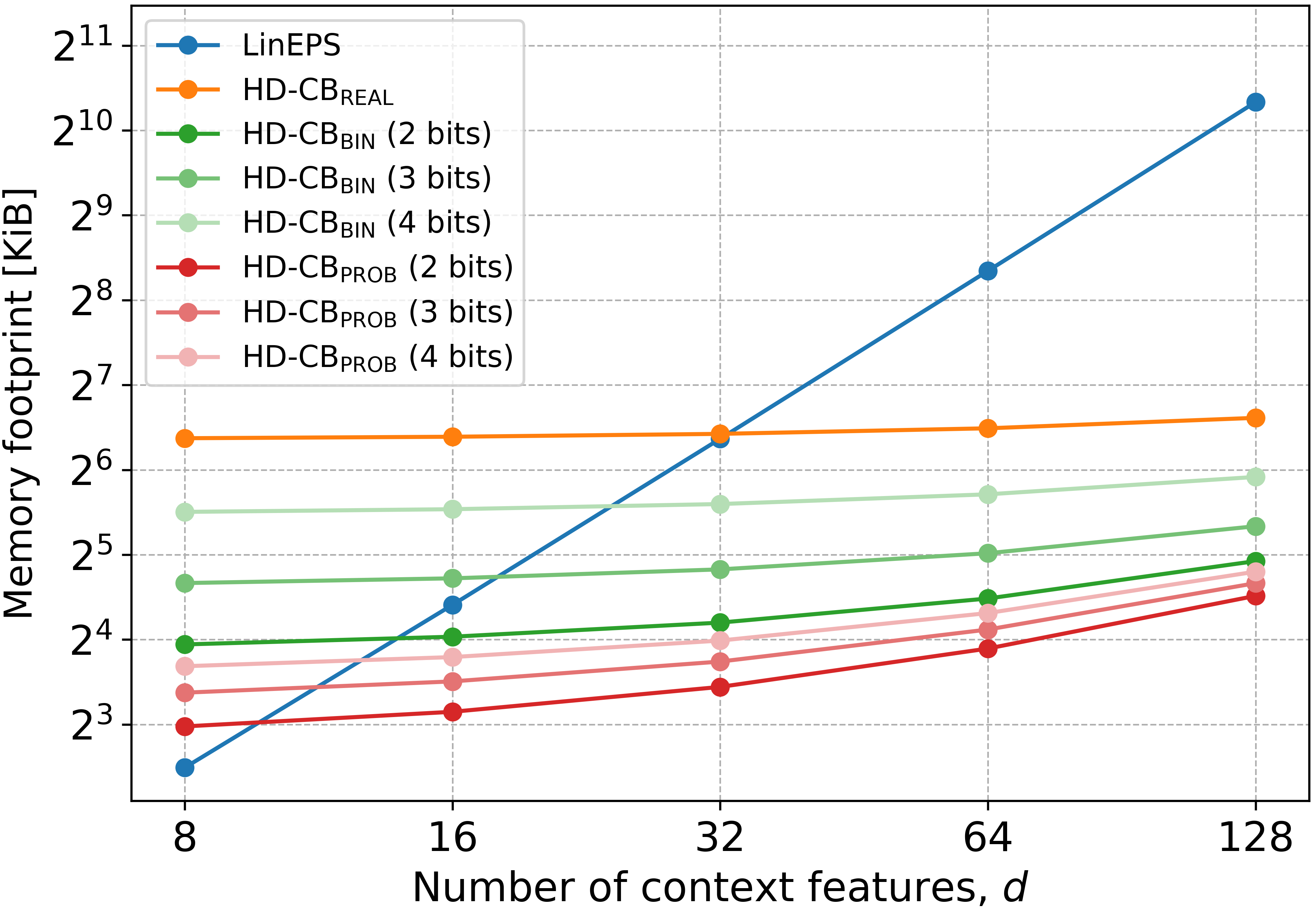}
    \caption{Memory footprint (KiB, log scale) vs. context dimensionality $d$. HD-CB\textsubscript{PROB} (red) consistently requires less memory than HD-CB\textsubscript{BIN} (green) at the same bitwidth as it does not require auxiliary counters or separate copies for inference.}
    \label{fig:memory_footprint}
\end{figure}

Figure~\ref{fig:memory_footprint} compares the memory footprint of the approaches as a function of the context feature dimension $d$ (reported in KiB on a logarithmic scale), for $d\in\{8,16,32,64,128\}$ and for 2-, 3-, and 4-bit instantiations of HD-CB\textsubscript{BIN} and HD-CB\textsubscript{PROB}. While LinEPS exhibits quadratic scaling with $d$, the HD-CB variants remain constant with respect to context dimensionality, scaling only with $D$. At any fixed bitwidth, HD-CB\textsubscript{PROB} requires \textit{strictly less memory} than HD-CB\textsubscript{BIN}. This efficiency stems from the architectural simplicity of the probabilistic approach. HD-CB\textsubscript{BIN} must maintain additional machinery per action: i) a $Q$-bit integer hypervectors to track history, ii) binarized copies for efficient action selection, and iii) per-action update counters to trigger re-binarization. In contrast, HD-CB\textsubscript{PROB} stores action hypervectors directly in their predefined bounded format ($\lceil \log_2(2\kappa+1)\rceil$ bits) and no additional copies or counters are required.

Overall, these results confirm that the proposed approach provides a better trade-off: compared to HD-CB\textsubscript{BIN}, it systematically yields a higher average reward at equal bitwidth, while simultaneously reducing memory footprint and avoiding the performance degradation induced by periodic binarization.

\section{Conclusion}\label{sec:conclusion}
As the demand for on-device intelligence continues to grow, enabling adaptive decision-making under strict resource constraints has become important. 
Standard linear contextual bandits have unfavorable quadratic scaling, while existing approaches based on hyperdimensional computing face a dilemma between growing memory footprint and information loss introduced by binarization.
This paper introduced \emph{probabilistic HD-CB}, a hardware-efficient approach that resolves this dilemma. 
By replacing deterministic accumulation with a saturating probabilistic update rule, the proposed approach incorporates new information through the \emph{frequency} of updates rather than the magnitude of stored values. 
This allows the agent to maintain high-precision rankings of available actions using only low-precision components.
Experimental evaluation on the Open Bandit Pipeline library demonstrates that the probabilistic HD-CB outperforms the binarized variant at equal bitwidth. 
Notably, a 3-bit implementation ($\kappa=3$) achieves cumulative rewards comparable to the real-valued variant, while requiring significantly less memory than even the binarized variant, as it does not use auxiliary counters.
Together, these results position probabilistic HD-CB as a compelling alternative for edge applications, offering a way to deploy adaptive, autonomous agents on devices with tight power, memory, and compute budgets. 
Future work includes evaluating the {probabilistic HD-CB} on real-world logged bandit datasets and under more complex reward distributions, as well as providing direct latency and energy measurements on embedded hardware to further substantiate the edge-deployment claims.

\bibliographystyle{IEEEtran} 
\balance
\bibliography{References_sh}

\end{document}